# Ferrograph image classification

Peng Peng*, Jiugen Wang

Faculty of Mechanical Engineering, Zhejiang University, Hangzhou, 310027, China

*Corresponding author: Peng Peng

E-mail: pengpzju@163.com

**Abstract**:

It has been challenging to identify ferrograph images with a small dataset and various scales of wear particle. A novel model is proposed in this study to cope with these challenging problems. For the problem of insufficient samples, we first proposed a data augmentation algorithm based on the permutation of image patches. Then, an auxiliary loss function of image patch permutation recognition was proposed to identify the image generated by the data augmentation algorithm. Moreover, we designed a feature extraction loss function to force the proposed model to extract more abundant features and to reduce redundant representations. As for the challenge of large change range of wear particle size, we proposed a multi-scale feature extraction block to obtain the multi-scale representations of wear particles. We carried out experiments on a ferrograph image dataset and a mini-CIFAR-10 dataset. Experimental results show that the proposed model can improve the accuracy of the two datasets by 9% and 20% respectively compared with the baseline.

**Keywords**: Ferrograph image classification, Multi-scale, Few-shot learning, Image patch permutation, Feature extraction loss

# 1 Introduction

The operation of most mechanical equipment is accompanied by mutual contact between parts, and the mutual contact will lead to wear of parts [1]. With the increase of machine service time, the wear between parts will gradually deteriorate and lead to equipment failure eventually. Therefore, it is of great significance to detect the wear of equipment regularly [2, 3].

Ferrograph is a feasible wear detection technology. It can determine the wear source, wear mechanism and wear severity of equipment by analyzing the products of wear [4]. Compared with vibration techniques and acoustic emission approaches, ferrograph technology presents wear products of equipment through ferrograph image. Hence, the fault signal obtained is more intuitive.

The early analysis of ferrograph image mainly relies on manual work, but the results of manual analysis method are subjective and time-consuming. In the 1990s, researchers proposed the automation method for ferrograph image recognition [5]. The automatic ferrograph image recognition methods can obtain the result quickly and objectively. Therefore, it has become one of the research hotspots of ferrograph. Myshki et al. proposed an artificial neural network to identify wear particles [6]. Wang et al. combined principal component analysis and gray correlation analysis methods to classify fatigue and sliding wear debris [7]. Peng et al. proposed a classification decision tree to recognize five kinds of wear particles [8]. Xu et al. designed a genetic programming evolution feature to determine wear debris [9]. The above methods of wear particle classification mainly focus on feature extraction of wear debris. However, the feature extraction relies on manual experience and the classification results depend on the quality of the extraction features. In recent years, deep learning has proven to be able to directly extract image features. Compared with traditional methods, deep learning methods have overwhelming advantages for image classification, detection and segmentation [10]. In the aspect of ferrograph image, researchers have also conducted a series of studies on ferrograph image processing with deep learning [11-17]. Peng et al. used transfer learning to solve the problem of overlapping wear particle classification [11]. Wang [16] et al. merged BP network and convolutional neural network to classify multiple wear particles. Zhang [15] et al. proposed a metric learning method to identify the unknown type of wear particles. Wang [13] et al. optimized the convolutional network to recognize fatigue wear particles and sliding wear particles. Peng [12] proposed the WP-DRnet network to detect the wear particles in ferrograph image.

Generally, compared with conventional machine learning methods, deep learning approaches can extract more robust features of wear particle and achieve a more accurate result on the classification of complex ferrograph images. It can accurately identify various wear particles even though the diverse background color and blur wear particle exist in ferrograph images [17].

However, the current deep learning methods also have the following shortcomings: (1) Abundant samples are required for training a deep learning model. Thus, a large number of experiments are needed to implement to obtain the specific types of fault samples. (2) Deep learning models usually obtain redundant features or have unimportant filters. (3) The performance of the deep learning method is often degraded when the object at different scales.

A new deep learning model is proposed to solve the above problems. The main contributions of this paper are as follows:

(1) A new data augmentation algorithm based on image patch permutation is proposed to cope with the problem of insufficient ferrograph images.

(2) An image patch permutation recognition loss function is proposed to constrain the search space of the model.

(3) A feature extraction loss function is proposed to improve the representation capability of the proposed model.

(4) We designed a multi-scale feature extraction block to obtain the different scales of wear particle features.

# 2 Method

As shown in Figure 1, the proposed model includes four parts: a data augmentation algorithm based on image patch permutation, a multi-scale feature extraction block, an image patch permutation recognition loss function and a feature extraction loss function. We adopt ResNet18 [10] as our baseline model in this study. The framework of the ResNet18 is shown in Figure 1(a).

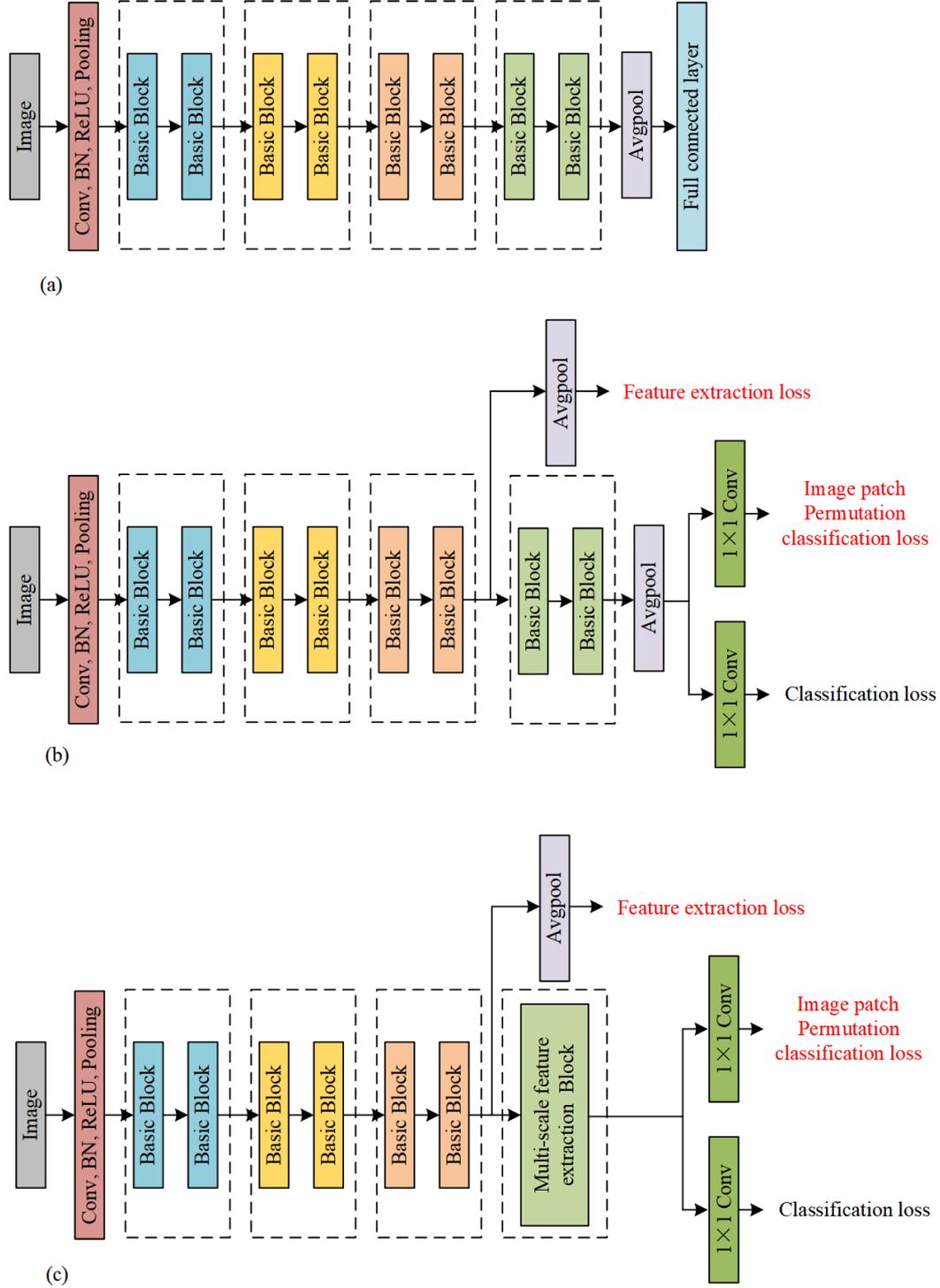

Figure 1. The framework of three different models. (a) ResNet18 [10], (b) ResNet18 with feature extraction loss and image patch permutation classification loss, (c) The improved ResNet18.

## 2.1 Data augmentation by image patch permutation

In terms of ferrograph, we need to conduct a large number of experiments to obtain enough samples. In order to reduce the dependence of model on data, data augmentation is a nice trick. Existing methods of data augmentation include data augmentation based on a single sample, data augmentation based on multi-sample [18-20], generative adversarial networks [21] and autoaugment [22]. Different from these approaches, we proposed a new data augmentation algorithm.

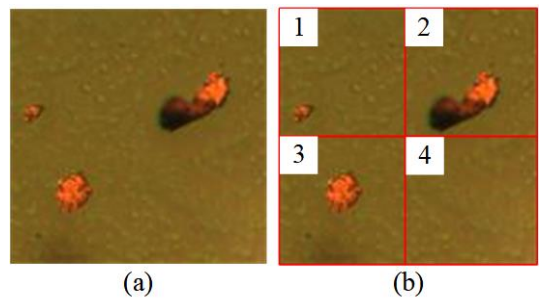

Figure 2. A ferrograph image and its patch allocation.

As shown in Figure 2, we first divide the original image into four patches. Then, we number each patch from left to right and top to bottom, as shown in Figure 2(b). In order to achieve the purpose of data augmentation, we rearrange the numbering order of the four image patches to obtain different permutations. According to the permutation formula, we have a total of 24 permutations. Hence, for each image, we can generate 24 samples. Figure 3 shows the results of data augmentation in Figure 2(a).

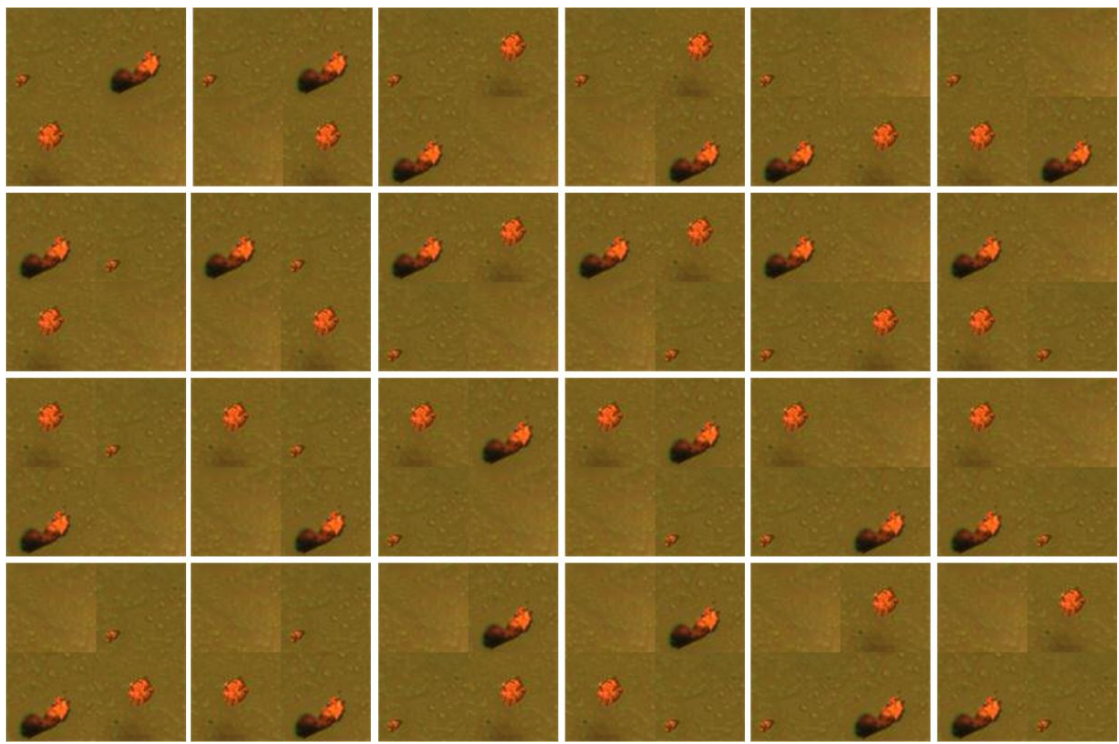

Figure 3. The generated images of Figure 2(a) after data augmentation.

## 2.2 Multi-scale feature extraction block

The performance of deep learning methods often deteriorated when an object with large scale variations. As shown in Figure 4, various scales of wear particles also exist in ferrograph images. Many studies, such as classification, detection and segmentation have fully proved that stuff with different scales could cause difficulty in prediction. Thereby, multi-scale representations are required for robust feature extraction. There are many approaches based on data or network to obtain multi-scale representations of an object. The methods based on data include data augmentation with multi-scale and image pyramids [23]. In terms of network, different pooling size, various convolution kernel size, feature pyramids and prediction based on different scales are the very practical tricks [24-28]. For simplicity, we proposed a new multi-scale feature extraction module in this study. The framework of the multi-scale feature extraction block is shown in Figure 5. We adopt multi-scale pooling operations to obtain different sizes of receptive fields, and we apply depth-wise separable convolution to reduce parameters of the block. All the 3×3 convolutions in Figure 5 are depth-wise separable convolutions.

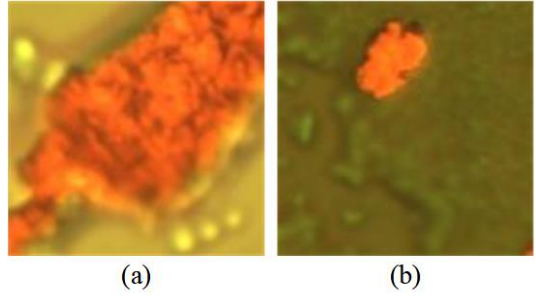

Figure 4. Wear particles with different scales.

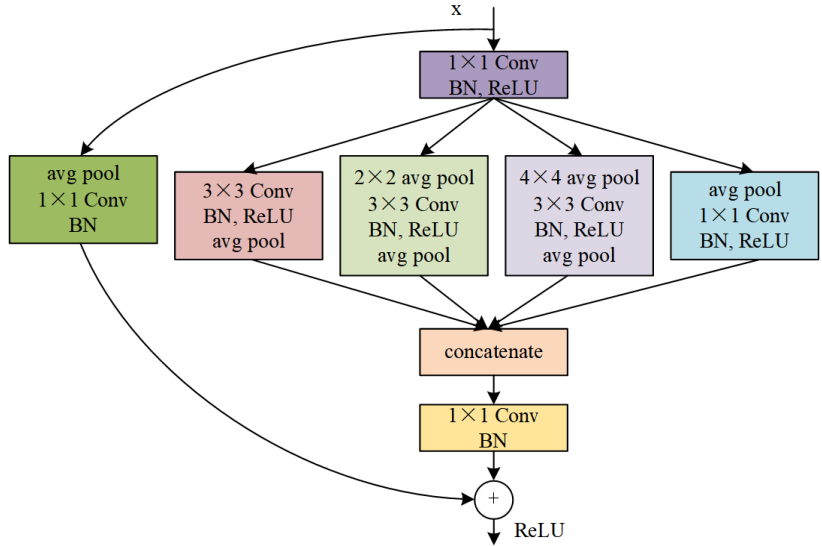

Figure 5. The proposed multi-scale feature extraction block.

## 2.3 Image patch permutation recovery loss

The deep learning model is easy overfitting on small datasets. Researches show that multi-task learning is a practicable way to deal with overfitting due to insufficient data [29]. In essence, multi-task learning is capable of reducing overfitting by constraining the hypothesis space of the network. Generally, the large model has a big hypothesis space which makes difficulties to find a nice hypothesis with strong generalization ability. Multi-task learning can constrain the search space of the model through the association between different tasks. Therefore, it is relatively easy to find a suitable hypothesis space with multi-task learning. In this research, we also adopt a multi-task learning strategy by adding an image patch permutation classification loss.

In section 2.1, we proposed an image patch permutation algorithm. For each image, we can obtain 24 permutation results. As shown in Figure 1, we design another head network to determine the permutation status of the current input image. The data augmentation method generates 24 kinds of images for each image. Therefore, the samples of each category are balanced. We employ the cross-entropy loss function to optimize this head network.

## 2.4 Feature extraction loss

Although deep convolutional neural networks are able to automatically extract image features through multiple consecutive convolutional layers, the phenomena of redundant representation and invalid kernel often occur [30]. Therefore, the trained larger model is often compressed, so that it can be applied on mobile devices without significant accuracy loss [31]. Model compress is not needed in the intelligent system of wear particle classification since it usually runs on a computer. However, in order to improve the representative ability of the model on small datasets, we need more constraints for the model.

The principal component analysis algorithm shows that if the variance of the data after coordinate transformation can be maximized, then the new data can retain the original data information to the greatest extent [32]. This inspires us that if the variance of the feature representation learned by the network is larger, can it be better to reduce the redundant representation? Based on this assumption, we propose the following feature extraction loss function:

$$Y_{B \times C} = \frac{1}{W \times H} \sum_{i=0}^{W} \sum_{j=0}^{H} X_{B \times C \times i \times j} \tag{1}$$

$$L_{std} = e^{-\sqrt{\frac{1}{C}\sum_{i=0}^{C}(Y_{B \times i} - \frac{1}{C}\sum_{j=0}^{C}Y_{B \times C})^2}} \tag{2}$$

Where $X$ is the input feature; $B$, $C$, $H$ and $W$ indicate the batch size, the channel numbers, the high and the width of $X$ respectively; $Y$ is the average of $X$ in the dimensions of height and width; $L_{std}$ is the feature extraction loss function which makes the $X$ have a large standard deviation.

As for the unimportant kernel, we propose another feature extraction loss function:

$$L_{mean} = e^{\frac{-1}{C \times W \times H}\sum_{k=0}^{C}\sum_{i=0}^{W}\sum_{j=0}^{H} X_{B \times i \times j \times k}} \tag{3}$$

Where $L_{mean}$ is another feature extraction loss function.

The loss function $L_{mean}$ encourages each feature map to obtain a larger activation value. Combined with the loss function $L_{std}$ we obtain the final feature extraction loss function as follows:

$$L_{feature} = 0.5(L_{std} + L_{mean}) \tag{4}$$

Where $L_{feature}$ is the final feature extraction loss function.

Combining the classification loss, image patch permutation loss and feature extraction loss, we can obtain the final loss function as follows:

$$L_{total} = L_{classification} + 0.5 L_{permutation} + L_{feature} \tag{5}$$

Where $L_{total}$ is the final loss function and $L_{permutation}$ is the image patch permutation loss.

# 3 Experiments and discussions

## 3.1 Classification of ferrograph images

We first collected oil samples from RV (rotate vector) reducer, bearing and gearbox. Then, ferrograph images were obtained from these oil samples by ferrograph. Finally, the images are randomly divided into a training set and test set. There are 909 images in the training set and 881 images in the test set. There are seven types of ferrograph images in the training set and test set, namely, background images, fatigue particles, oxide particles, sphere particle, fatigue particles and oxide particle, fatigue particles and sphere particle, oxide particle and sphere particle.

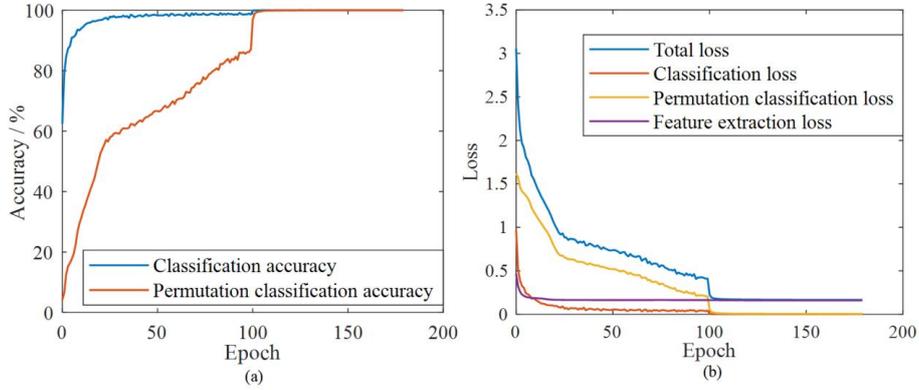

Figure 6. Accuracy and loss of ResNet18 on ferrograph image dataset

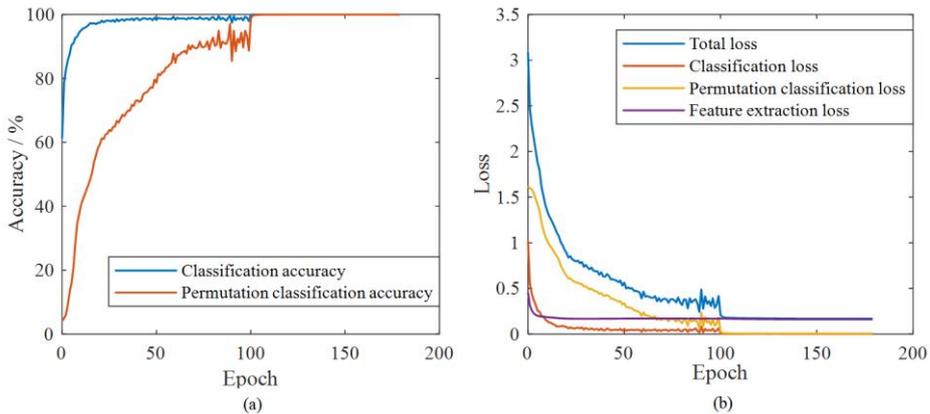

Figure 7. Accuracy and loss of improved ResNet18 on ferrograph image dataset.

We apply the proposed model in this dataset. The batch size is set into 64, the optimizer is the stochastic gradient descent method with an initial learning rate of 0.1. The learning rate is divided by 10 at 100 and 140epochs. The curve of accuracy and loss function in training processing is shown in Figure 6 and Figure 7. The experimental results are shown in Figure 8 and Figure 9.

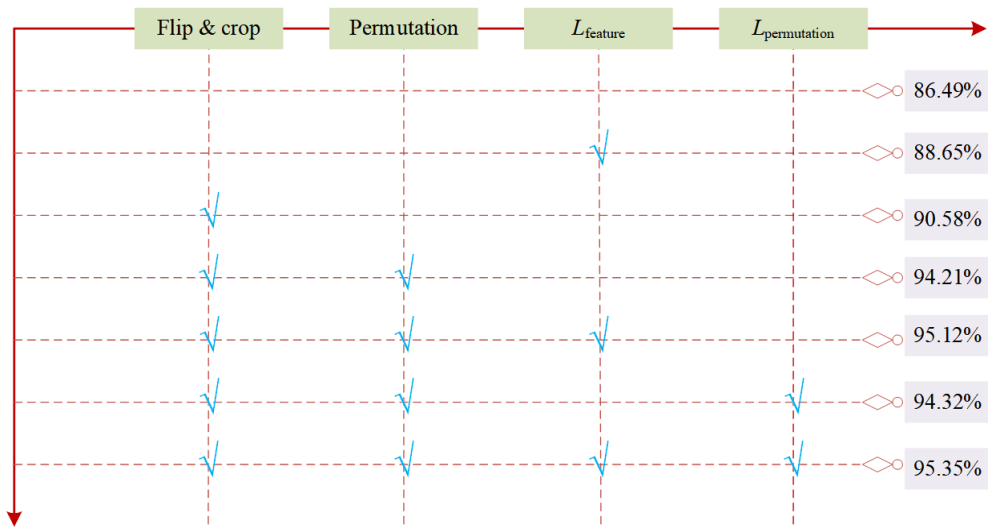

Figure 8. Experimental results of ResNet18 and proposed loss function on ferrograph image dataset.

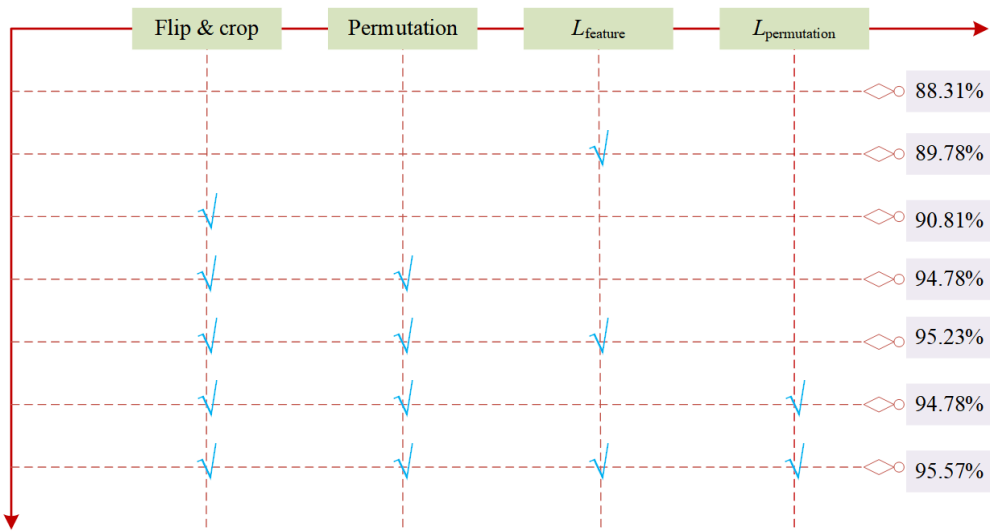

Figure 9. Experimental results of improved ResNet18 and proposed loss function on ferrograph image dataset.

## 3.2 Experimental results on mini-CIFAR-10

The CIFAR-10 dataset consists of 50k training images and 10k testing images in 10 classes [33]. The training images of the CIFAR-10 dataset are far more than the training ferrograph images. In order to prove the performance of the proposed model on small datasets, we randomly collected 100 images from each class to form a mini dataset. We call mini-CIFAR-10 in this study. Hence, we have 1000 training images in the mini-CIFAR-10 dataset and the test set of the mini-CIFAR-10 is the same with the original CIFAR-10 test set. Experiments are conducted on the mini-CIFAR-10 dataset, and the results are shown in Figure 10 and Figure 11.

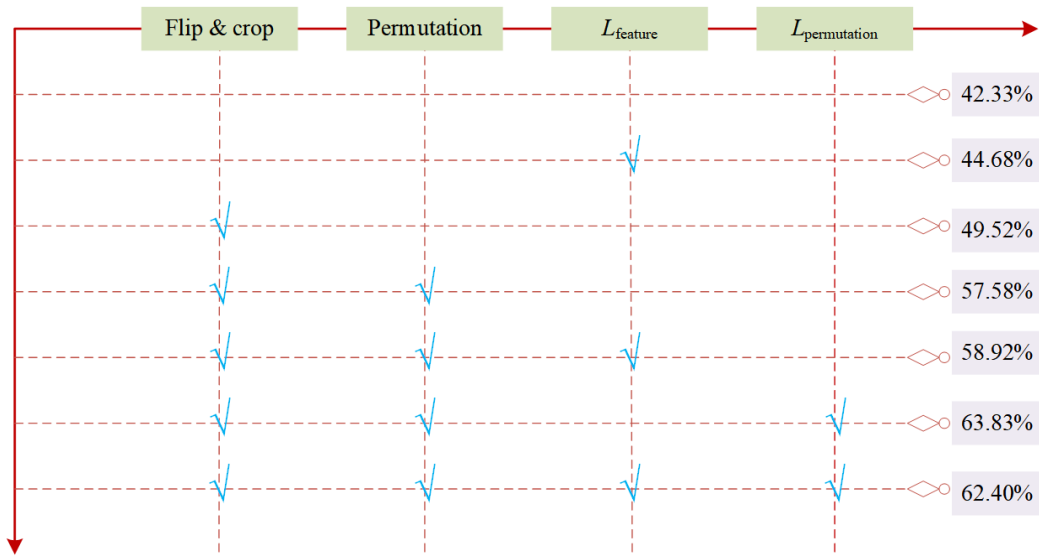

Figure 10. Experimental results of ResNet18 and proposed loss function on mini-CIFAR-10.

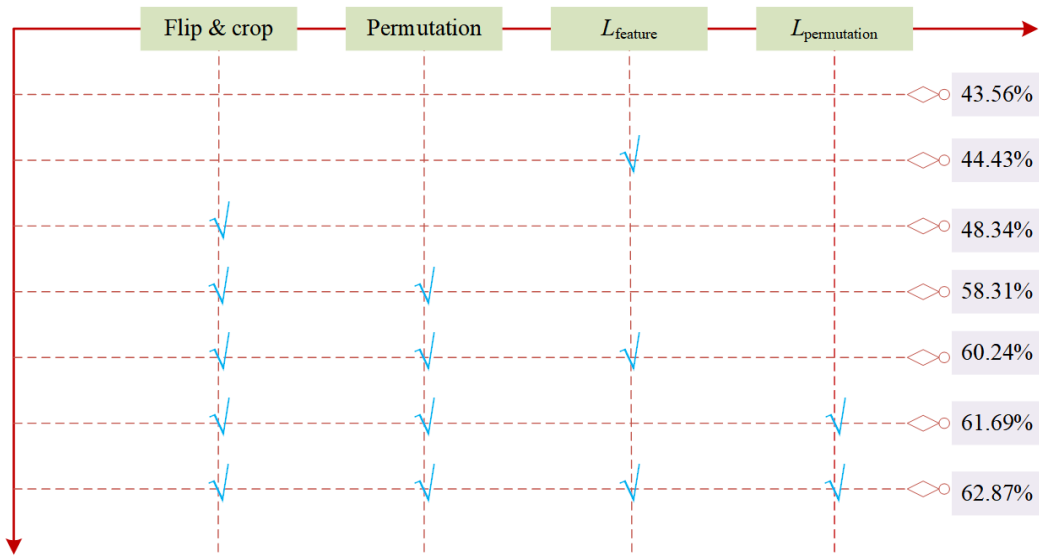

Figure 11. Experimental results of improved ResNet18 and proposed loss function on mini-CIFAR-10.

## 3.3 Discussion

The effects of the four parts of the proposed models on the two datasets will be analyzed in this section. As shown in Figure 8-11, the data augmentation algorithm improves the performance of both two dataset and two models. Besides, the proposed multi-scale feature extraction block also works on both two dataset and two models, especially for dataset without data augmentation. The main reason is that the number of parameters in the improved ResNet18 is less than that in the ResNet18. Thus, the overfitting of the improved ResNet18 is alleviated.

The image patch recovery loss improves the performance on both two datasets. However, it obtains a better result on the mini-CIFAR-10 dataset and it improves the performance on the ferrograph image dataset slightly. The main reason may be that most of the wear particles are not located in the centre of the ferrograph image, while most of the targets of the mini- CIFAR-10 dataset are present in the centre of the image. Therefore, the targets in the mini- CIFAR-10 dataset are more likely to be evenly divided into four image patches when the proposed data augmentation algorithm applied on an image. Hence, the network can determine the permutation type easily by the shape of the objects. However, most of the wear particles are not displayed in the centre of a ferrograph image. Thus, wear particles are more likely to fall into one image patch, as shown in Figure 3. Hence, it is difficult for the network to identify the permutation mode of the input image patches.

The proposed feature extraction loss also improves the performance on both datasets and models. In order to show the effects of the feature extraction loss on the ResNet18 and the improved ResNet18, we feed Figure 2(a) into the two models and visualize the feature maps of layer3 for both models. The results are shown in Figure 12. It can be found in Figure 12, the model with feature extraction loss makes a bigger difference between each feature map and most of the feature maps are activated.

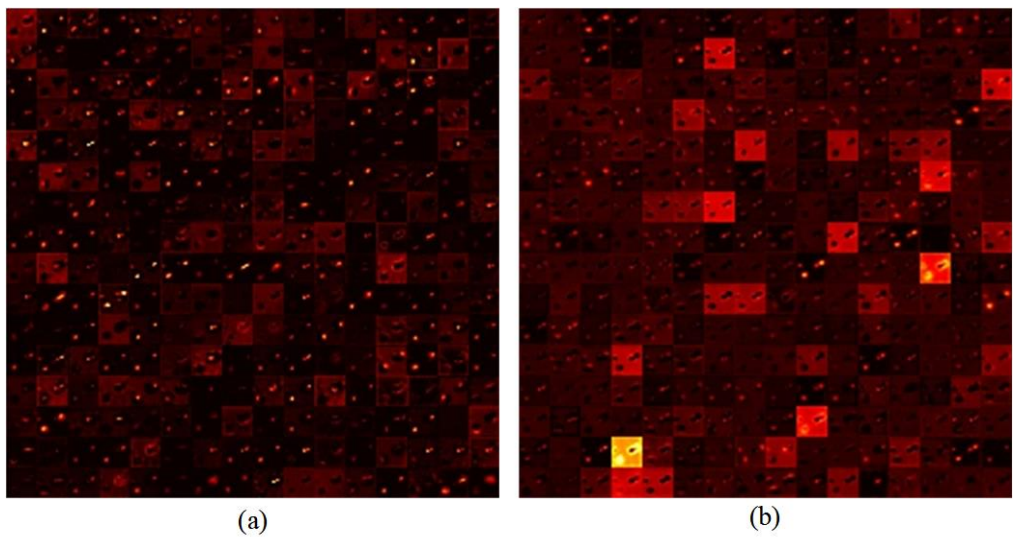

Figure 12. Output feature maps of layer3. (a) Without feature extraction loss, (b) With feature extraction loss.

# 4 Conclusion

(1) A new model is proposed in this study to solve the problem of wear particle classification with insufficient samples and multi-scale objects.
(2) The proposed data augmentation algorithm improves the performance of the model on both the ferrograph image dataset and the mini-CIFAR-10 dataset. It fully reflects the importance of data augmentation for small datasets. Besides, the multi-scale feature extraction module can also improve the performance of the model.
(3) The image patch permutation classification loss also works on both dataset, especially for the mini-CIFAR-10 dataset where the most of targets are presented in the centre of the images.
(4) The feature extraction loss can make the variance of the features larger and reduce feature redundancy. Moreover, unimportant features in the model with feature extraction loss are less than the model without the feature extraction loss.

# Reference

[1] Meng Y G, Xu J, Jin Z M, et al. A review of recent advances in tribology [J]. Friction, 2020, 8(2): 221-300.

[2] Wakiru J M, Pintelon L, Muchiri P N, et al. A review on lubricant condition monitoring information analysis for maintenance decision support [J]. Mechanical systems and signal processing, 2019, 118: 108-132.

[3] Feng P, Borghesani P, Smith W A, et al. A review on the relationships between acoustic emission, friction and wear in mechanical systems [J]. Applied Mechanics Reviews, 2020, 72(2).

[4] Wu T H, Mao J H, Wang J T, et al. A new on-line visual ferrograph[J]. Tribology transactions, 2009, 52(5): 623-631.

[5] Peng Z, Kirk T B. Wear particle classification in a fuzzy grey system [J]. Wear, 1999, 225: 1238-1247.

[6] Myshkin N K, Kwon O K, Grigoriev A Y, et al. Classification of wear debris using a neural network [J]. Wear, 1997, 203: 658-662.

[7] Wang J, Wang X. A wear particle identification method by combining principal component analysis and grey relational analysis [J]. Wear, 2013, 304(1-2): 96-102.

[8] Peng Y, Wu T, Cao G, et al. A hybrid search-tree discriminant technique for multivariate wear debris classification [J]. Wear, 2017, 392: 152-158.

[9] Xu B, Wen G, Zhang Z, et al. Wear particle classification using genetic programming evolved features [J]. Lubrication Science, 2018, 30(5): 229-246.

[10] He K, Zhang X, Ren S, et al. Deep residual learning for image recognition [C]. Proceedings of the IEEE conference on computer vision and pattern recognition. 2016: 770-778.

[11] Peng P, Wang J. Wear particle classification considering particle overlapping [J]. Wear, 2019, 422: 119-127.

[12] Peng Y, Cai J, Wu T, et al. WP-DRnet: A novel wear particle detection and recognition network for automatic ferrograph image analysis [J]. Tribology International, 2020: 106379.

[13] Wang S, Wu T, Zheng P, et al. Optimized CNN model for identifying similar 3D wear particles in few samples[J]. Wear, 2020: 203477.

[14] Wang J, Liu X, Wu M, et al. Direct detection of wear conditions by classification of ferrograph images [J]. Journal of the Brazilian Society of Mechanical Sciences and Engineering, 2020, 42(4): 1-10.

[15] Zhang T, Hu J, Fan S, et al. CDCNN: A Model Based on Class Center Vectors and Distance Comparison for Wear Particle Recognition [J]. IEEE Access, 2020, 8: 113262-113270.

[16] Wang S, Wu T H, Shao T, et al. Integrated model of BP neural network and CNN algorithm for automatic wear debris classification [J]. Wear, 2019, 426: 1761-1770.

[17] Peng P, Wang J. FECNN: A promising model for wear particle recognition [J]. Wear, 2019, 432: 202968.

[18] Inoue H. Data augmentation by pairing samples for images classification. arXiv preprint arXiv:1801.02929, 2018.

[19] Zhang H, Cisse M, Dauphin Y N, et al. mixup: Beyond empirical risk minimization. arXiv preprint arXiv:1710.09412, 2017.

[20] Chawla N V, Bowyer K W, Hall L O, et al. SMOTE: synthetic minority over-sampling technique

[J]. Journal of artificial intelligence research, 2002, 16: 321-357.

[21] Goodfellow I, Pouget-Abadie J, Mirza M, et al. Generative adversarial nets[C]. Advances in neural information processing systems. 2014: 2672-2680.

[22] Cubuk E D, Zoph B, Mane D, et al. Autoaugment: Learning augmentation policies from data [J]. arXiv preprint arXiv:1805.09501, 2018.

[23] Liu Z, Gao G, Sun L, et al. IPG-Net: Image Pyramid Guidance Network for Small Object Detection[C]. Proceedings of the IEEE/CVF Conference on Computer Vision and Pattern Recognition Workshops. 2020: 1026-1027.

[24] Zhao H, Shi J, Qi X, et al. Pyramid scene parsing network[C]. Proceedings of the IEEE conference on computer vision and pattern recognition. 2017: 2881-2890.

[25] Chen L, Papandreou G, Kokkinos I, et al. DeepLab: Semantic Image Segmentation with Deep Convolutional Nets, Atrous Convolution, and Fully Connected CRFs [J]. IEEE Transactions on Pattern Analysis and Machine Intelligence, 2018, 40(4):834-848.

[26] Lin T Y, Dollár P, Girshick R, et al. Feature pyramid networks for object detection[C]. Proceedings of the IEEE conference on computer vision and pattern recognition. 2017: 2117-2125.

[27] He J, Deng Z, Qiao Y. Dynamic multi-scale filters for semantic segmentation[C]. Proceedings of the IEEE International Conference on Computer Vision. 2019: 3562-3572.

[28] Liu W, Anguelov D, Erhan D, et al. Ssd: Single shot multibox detector[C]. European conference on computer vision. Springer, Cham, 2016: 21-37.

[29] Zhang Y, Yang Q. A survey on multi-task learning [J]. arXiv preprint arXiv:1707.08114, 2017.

[30] Meng F, Cheng H, Li K, et al. Filter grafting for deep neural networks[C]. Proceedings of the IEEE/CVF Conference on Computer Vision and Pattern Recognition. 2020: 6599-6607.

[31] Xu G, Liu Z, Li X, et al. Knowledge Distillation Meets Self-Supervision. arXiv preprint arXiv:2006.07114, 2020.

[32] Wold S, Esbensen K, Geladi P. Principal component analysis [J]. Chemometrics and intelligent laboratory systems, 1987, 2(1-3): 37-52.

[33] Krizhevsky A, Hinton G. Learning multiple layers of features from tiny images [R]. 2009.